\pdfoutput=1

\documentclass[11pt]{article}

\usepackage{ACL2023}

\usepackage{times}
\usepackage{latexsym}
\usepackage{pgfplots}
\usepgfplotslibrary{colorbrewer}
\pgfplotsset{compat=1.18}
\usepackage[T1]{fontenc}

\usepackage[utf8]{inputenc}

\usepackage{microtype}

\usepackage{inconsolata}
\usepackage{graphicx}
\usepackage{multirow}

\title{Leveraging Annotator Disagreement for Text Classification}

\author{Jin Xu \\
  University of Twente\\
  \texttt{jinxu130817@gmail.com} \\\And
 Mariët Theune \\
  University of Twente\\
  Human-Media Interaction\\
  \texttt{m.theune@utwente.nl} \\\And
  Daniel Braun \\
  University of Twente\\
 Industrial Engineering and\\ 
Business Information Systems\\
  \texttt{d.braun@utwente.nl}}

\begin{document}
\maketitle
\begin{abstract}
It is common practice in text classification to only use one majority label for model training even if a dataset has been annotated by multiple annotators. Doing so can remove valuable nuances and diverse perspectives inherent in the annotators’ assessments. This paper proposes and compares three different strategies to leverage annotator disagreement for text classification: a probability-based multi-label method, an ensemble system, and instruction tuning. All three approaches are evaluated on the tasks of hate speech and abusive conversation detection, which inherently entail a high degree of subjectivity. Moreover, to evaluate the effectiveness of embracing annotation disagreements for model training, we conduct an online survey that compares the performance of the multi-label model against a baseline model, which is trained with the majority label.

The results show that in hate speech detection, the multi-label method outperforms the other two approaches, while in abusive conversation detection, instruction tuning achieves the best performance. The results of the survey also show that the outputs from the multi-label models are considered a better representation of the texts than the single-label model.


\end{abstract}

\section{Introduction}

Employing multiple annotators for data annotation and afterwards using the majority annotation for model training is a widely adopted practice to mitigate biases and allow for error detection and correction \cite{sabou_corpus_2014}. 
However, such procedures also remove genuine disagreement between annotators that can provide valuable insights, e.g. for subjective tasks like detection of hate speech, emotions, or sexism, but also for more objective tasks like legal or medical decision making. 
In recent years, the practice of only considering majority annotations has been increasingly criticized and many researchers have started to advocate for better ways to deal with disagreement between annotators \citep{basile_toward_2021, uma_learning_2021, plank-2022-problem, braun2023beg}.

In this article, we propose three different strategies to leverage annotator disagreement during the training of text classification models: a probability-based multi-label approach, an ensemble system approach, and an instruction tuning approach. We compare these strategies against a baseline model that is trained on the majority labels derived from the multiple annotations. We choose two text classification tasks which inherently entail a high degree of subjectivity for the evaluation: hate speech detection and abuse detection in conversations. In our chosen datasets \citep{toraman_large-scale_2022, cercas-curry-etal-2021-convabuse}, these two tasks exhibit different complexity and difficulty in terms of the label space: while the hate speech detection dataset contains binary labels, the abusive conversation detection dataset is not only annotated with abusive / non-abusive but also the severity of the abuse.

Our first approach tackles the tasks as a probability-based multi-label text classification problem. Instead of predicting specific labels to one instance, the model provides a probability distribution. The second approach imitates the process of annotation from multiple annotators with an ensemble system. The ensemble system consists of many sub-models, each of which is trained on different labels to capture the diverse viewpoints embedded in the annotations. Thirdly, we use instruction tuning. Specifically, we use a pre-trained generative model and inject explicit guidance into the training process to customize the model’s behavior. The performance of the proposed models is compared using cross entropy. To evaluate the effectiveness of incorporating multiple labels, we also conduct an online survey. This survey aims to investigate human preferences between the outputs generated by the multi-label model and a baseline model. 

The results show that on the hate speech dataset, the multi-label method outperforms the ensemble system and instruction tuning. Conversely, instruction tuning is the best-performing method on the abusive conversation dataset. Through multinomial test, the outputs from the multi-label model are considered more reasonable than those from the baseline model to characterize samples from the online survey. This proves the effectiveness of leveraging annotation disagreements for model training.





\section{Related Work}
\subsection{Sources of Disagreement}
Disagreement in annotations can originate from different sources. Natural language can be inherently complex and interpreted in multiple ways within a given context \cite{poesio_ambiguity_2020}. There are many subjective elements which may add an additional layer of intricacy to the understanding of texts, such as sentiments, opinions or nuanced expressions. Therefore, it is common that there are divergent interpretations among annotators. Furthermore, some sentences and even the definition of labels may contain vague or ambiguous statements \cite{russell_labelme_2008}, making it challenging for annotators to reach an agreement.

However, annotators themselves and their background can also have significant impact on the annotation results \cite{davani_dealing_2022}. Through post-annotation interviews, \citet{patton_annotating_2019}, for example, showed that annotators who come from communities discussed in gang-related tweets are more likely to rely on their lived experiences in the process of annotating when compared to graduate student researchers. This divergence results in distinct label judgments.  \citet{luo_desmog_2020} found that the political affiliation of annotators can significantly shape how they assess and annotate political stances.

\subsection{Handling Disagreement}
Majority voting involves aggregating annotations by selecting the label that the majority of annotators agree upon. Majority voting is easy to understand and implement and tends to perform well when the annotators share unanimous perspectives \cite{uma_learning_2021}. However, the employment of a majority voting method in annotation processes can unintentionally obscure nuanced viewpoints, especially for groups that are underrepresented in annotator pools \cite{prabhakaran-etal-2021-releasing}. 
To address this concern, it is important to ensure a diverse representation among annotators to foster a more comprehensive understanding of various perspectives, particularly those from underrepresented demographics \cite{wan_everyones_2023}.

Some studies have introduced alternative methods to majority voting in order to incorporate annotator disagreement in model training. \citet{chou_every_2019} modelled the label uncertainty and annotator idiosyncrasy simultaneously by using both hard label (majority voting) and soft label (the distribution of annotations). The results showed that the soft label contains useful information that significantly boosts the model performance. \citet{fornaciari_beyond_2021} proposed a multi-task neural network that was trained on soft label distribution over annotator labels. By integrating a divergence measurement between soft label and “true” label vector into the loss functions, they effectively mitigated overfitting and therefore improved model performance. \citet{davani_dealing_2022} introduced multi-annotator models where each annotator’s judgements were regarded as independent sub-task with a shared common representation of the annotation task. This approach enables to preserve and model the internal consistency in each annotator’s label. It also incorporates the systematic disagreements with other annotators. Similarly, the network architecture introduced by \citet{guan_who_2018} individually models annotation experts. In this approach, each expert’s model weight is calculated independently, and these individual weights are then averaged to facilitate ensemble recognition. To include the knowledge from annotators, \citet{fayek_modeling_2016} employed neural networks to build an ensemble system that consists of many models, with each model representing one annotator. Then the final results are obtained by combining the individual model outputs.

Although the approaches outlined above have improved the performance by leveraging annotation disagreements, they remained limited to identifying the majority label. The outputs, in the form of ``soft labels'' (probability distribution over labels), were still aggregated to single labels as final predictions. There is limited research focusing on evaluating the effectiveness of embracing multiple labels.


\section{Datasets}

In this section, the two datasets that have been used in this study will be briefly introduced.

\subsection{Hate Speech}

The first dataset is the “Large-Scale Hate Speech Dataset”\footnote{\url{https://github.com/avaapm/hatespeech/tree/master/dataset_v1}} published by \citet{toraman_large-scale_2022}. It consists of a total of 100,000 tweets (7,000 training, 1,500 validation, and 1,500 testing). Each tweet in the dataset is annotated by five annotators that have been selected randomly from a panel of 20 annotators. According to the annotation guidelines utilized by \citet{sharma-etal-2018-degree}, tweets are categorized as “Hate” if they target, incite violence against, threaten, or advocate for physical harm towards an individual or a group of people based on identifiable trait or characteristic. If tweets humiliate, taunt, discriminate against, or insult an individual or a group of people, they are annotated as “Offensive”. In the absence of these criteria, the tweets are labeled as “Normal” .

\subsection{Abuse Conversation}

The second dataset is the “Abuse in Conversational AI” dataset\footnote{\url{https://github.com/amandacurry/convabuse/tree/main}} (hereinafter referred to as “abusive conversation dataset”) published by \citet{cercas-curry-etal-2021-convabuse}. The data was collected from conversations between users and conversational AI systems, and consists of 2501 samples as training data, 831 as validation data and 853 as testing data. The data was annotated using an unbalanced rating scale proposed by  \citet{poletto_annotating_2019}, in which inputs are labelled on a scale from Not abusive, Ambiguous, Mildly abusive, Strongly abusive to Very strongly abusive. This annotation scheme offers insights into not only the presence of abusive content, but also the severity of the abuse.
In the annotation process, eight annotators were recruited, and each example is annotated by a minimum of three annotators. 

\section{Methodology}

\subsection{Baseline model}

The baseline model for this study is trained on the “ground truth” label that is aggregated via majority voting. Given BERT’s \citep{devlin-etal-2019-bert} notable performance in contextual understanding, we choose it as the pre-trained model. Since the baseline model outputs a single label, we augment its architecture by adding a fully connected layer to the last hidden state, thereby adapting the model structure to the specific prediction task.



\subsection{Probability-based multi-label method}
\begin{figure}
  \includegraphics[width=\columnwidth]{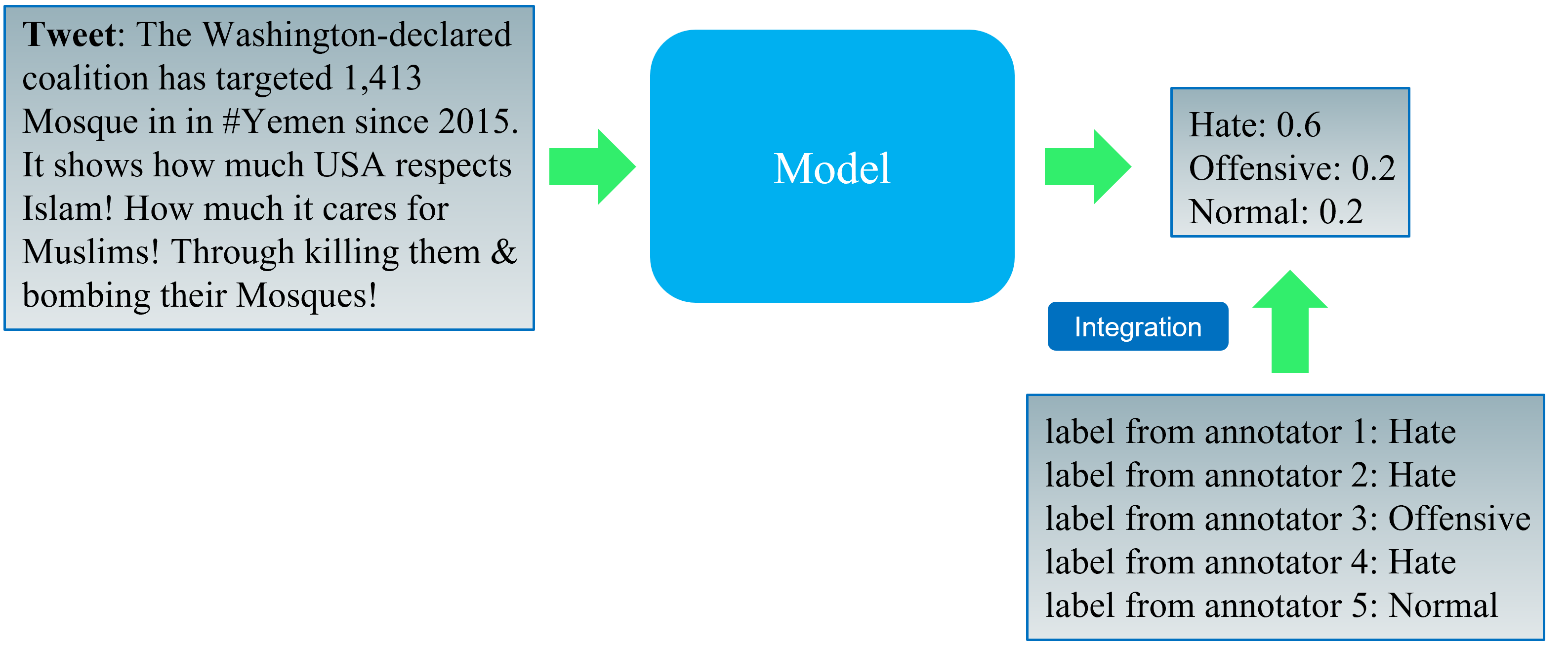}
  \caption{The framework of model training within the probability-based multi-label method.}
  \label{fig:multilabel}
\end{figure}

The task of identifying hate speech or abusive conversation can be regarded as a multi-label text classification problem, where a given piece of text can be associated with one or multiple labels simultaneously.
Unlike the traditional approaches that assign one or several exclusive labels to the input text \cite{jiang_identifying_2020}, our model predicts the probability of each label being associated with the given text. The approach is illustrated in Figure~\ref{fig:multilabel}. The model is trained on the probability distribution across different labels which is derived from individual annotations. Like the baseline model, the multi-label model also is based on BERT but fine-tuned with different types of target labels.

\subsection{Ensemble system}
\begin{figure*}
\center
  \includegraphics[width=.8\textwidth]{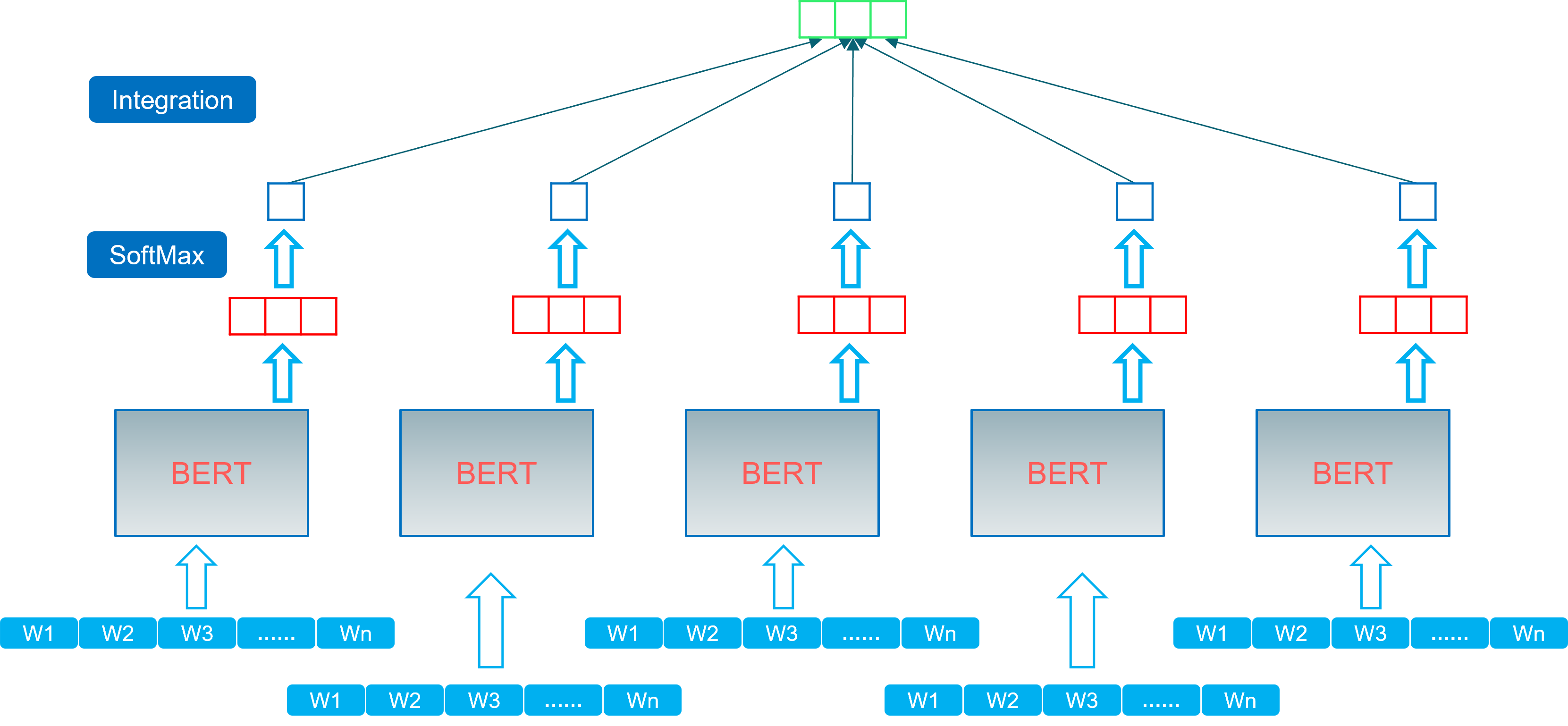}
  \caption{Fine-tuning BERT individually as sub-models within the ensemble system.}
  \label{fig:ensemble}
\end{figure*}

In the annotation process, multiple labels are assigned by different annotators. Inspired by this process, we propose an ensemble system consisting of several sub-models. 
As shown in Figure~\ref{fig:ensemble}, each sub-model is based on a BERT model that is fine-tuned individually on its respective set of labels. For each sub-model, the input is the text from one sample and the output is a multidimensional vector where each dimension corresponds to one category. After that, this vector is transformed by the SoftMax function and the dimension with highest probability is identified as the output of the sub-model. Finally, the predictions from all sub-models are combined and converted into a probability distribution of three- or five-dimensional vector.

In the abusive conversation dataset, the annotators assigned for each sample are clearly specified and identifiable. Therefore, within the ensemble system, each sub-model represents one specific annotator and is trained on that annotator’s labels. By contrast, in the hate speech dataset, each sample is labeled by five anonymous annotators. Despite the anonymity, training a model with such labels can potentially increase the robustness of sub-models since it helps to reduce the biases or inconsistencies introduced by individual annotators \cite{frenay_classification_2014}. Furthermore, the resulting labels are likely to reflect a diverse range of perspectives and interpretations of the data. Training sub-models on these diverse annotations can capture the variability in annotator judgments and enhance the model’s ability to generalize across different viewpoints \cite{audhkhasi_globally-variant_2013}.
Since the sub-models can show varying performances in the training and validation processes, typically,  the top n (n$\geq$3) best-performing sub-models are chosen to determine the final output. The ranking is based on their accuracy on the validation data.

\subsection{Instruction tuning}
\begin{figure}
  \includegraphics[width=\columnwidth]{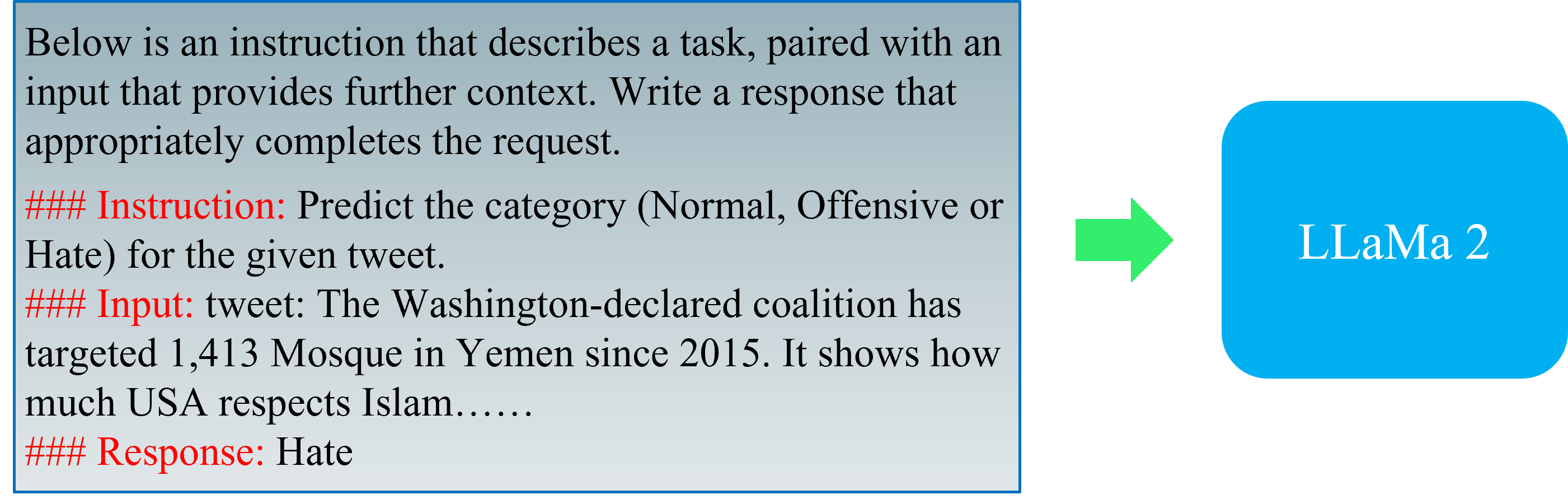}
  \caption{Fine-tuning LLaMa 2 as a sub-model with instruction tuning in the hate speech dataset.}
  \label{fig:hatespeech}
\end{figure}

\begin{figure}
  \includegraphics[width=\columnwidth]{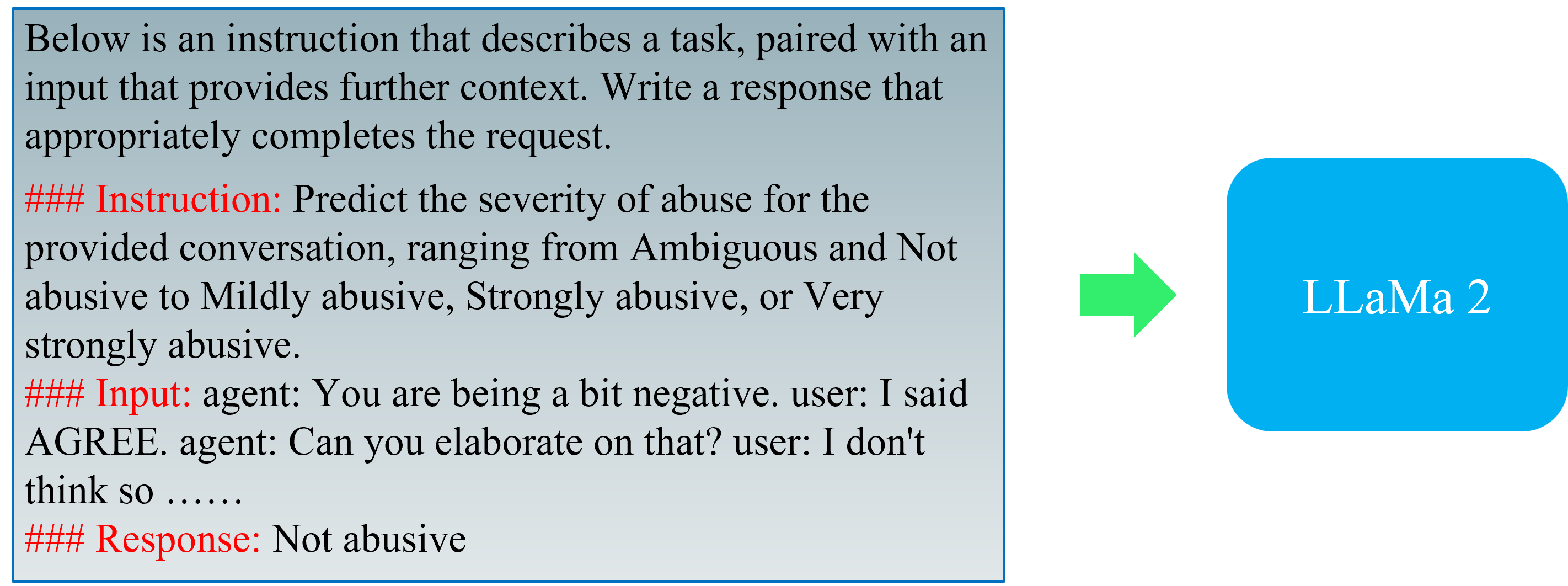}
  \caption{Fine-tuning LLaMa 2 as a sub-model with instruction tuning in the abusive conversation dataset.}
  \label{fig:abuse}
\end{figure}

Instruction tuning is the process of fine-tuning LLMs in a supervised fashion on a dataset consisting of pairs of instructions and outputs. The key idea is to provide the model with explicit instructions to enhance its performance and align it with specific objectives. Unlike traditional training approaches where models learn from data alone, instruction tuning injects explicit guidance into the training process. This approach allows for explicit customization of the model’s behavior. In this study, we ask the model to predict the class of hate speech or abusive conversation based on the input we construct. The input contains the task description, the instruction, the original text, and the annotation from one specific annotator (i.e. \textit{not} the majority label).
The approach of fine-tuning LLaMa 2 via instruction tuning on the two datasets is presented in Figure~\ref{fig:hatespeech} and Figure~\ref{fig:abuse}. On the left sides of the figures are the inputs fed into the pre-trained model. The input comprises the following four components: scenario description, instruction, text input and response.

Like the ensemble system, top n (n$\geq$3) best-performing sub-models are selected to contribute to the final predictions.

\subsection{Evaluation}


\subsubsection{Cross entropy}
The final output of our proposed models is a probability distribution across different labels, and in this scenario, a single “ground truth” label is no longer applicable for model evaluation. Instead, we use cross entropy to compare the distribution of annotations with model output. Cross entropy is one kind of statistical distance which measures how a probability distribution is different from a reference probability distribution. In the field of NLP, it has been used to quantify how well the model’s predicted distribution matches the annotation distribution \cite{pavlick_inherent_2019}. 


\subsubsection{Online survey}\label{sec:onlinesurvey}
Using cross entropy to evaluate the effectiveness of training models with multiple labels against models that only rely on the majority label is impossible due to the format disparity between the outputs generated. To bridge this gap, we conduct an online survey where participants specify their preference between annotations generated from the probability-based multi-label model and the baseline model. For each dataset, we select 10 samples, each featuring two annotations. Both annotations are in the form of probability distributions across different labels. One is generated from the baseline model trained with majority labels, which is, however, used to generate a probability distribution in the phase of inference. The other one is from the probability-based multi-label model. This model has the same structure as the baseline model and their only difference is the labels they were trained on. For each sample, participants are required to indicate which annotation they find is more reasonable to characterize the tweet or the abusive conversation.

\section{Results}


\subsection{Multi-Label Method}
\begin{table}
  \centering
  \begin{tabular}{cccc}
    \hline
    \textbf{} & \textbf{Training} & \textbf{Validation} & \textbf{Testing}\\
    \hline
    Hate speech    & 0.7613  & 0.7569 & 0.7638\\
    \begin{tabular}[c]{@{}c@{}}Abusive\\conversation\end{tabular} & 0.8861 & 0.9680 & 0.9834\\\hline
  \end{tabular}
  \caption{The average cross entropy of the probability-based multi-label model on two tasks.}
  \label{tab:multi_label_cross_entropy}
\end{table}

Table~\ref{tab:multi_label_cross_entropy} show the performance of the multi-label model on the two datasets. In this approach, the model demonstrates superior performance on the hate speech dataset compared to the abusive conversation dataset. In particular, the cross entropy for the hate speech dataset is 0.7638, while this value for the abusive conversation dataset is 0.9834. With a parameter size of 110 million, the multi-label model benefits from the extensive training data in hate speech dataset to optimize and align itself with the downstream task. In contrast, there are only 2501 training samples in the abusive conversation dataset, which can easily lead to overfitting in the process of training. The multi-label model exhibits relatively consistent losses across training, validation, and testing data in the hate speech dataset, indicating a good fit without signs of underfitting or overfitting. By comparison, in the abusive conversation dataset, losses during validation and testing are noticeably higher than during training. When the model encounters unseen data in validation and testing phases, the loss can be relatively high due to the lack of generalization.

\subsection{Ensemble system}

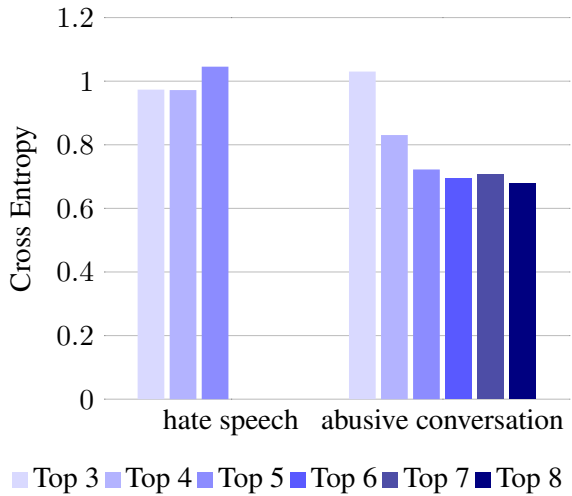
\begin{figure}
    \centering
    \begin{tikzpicture}
  \begin{axis}[
  width  = \linewidth,
  ybar,
    ylabel = Cross Entropy,
    x axis line style = { opacity = 0 },
    tickwidth         = 0pt,
    enlarge x limits=0.60,
    symbolic x coords = {hate speech, abusive conversation},
    xtick = data,
    ymin = 0,
    ymax = 1.2,
    ymajorgrids = true,
    legend style={at={(0.4,-0.15)},anchor=north,draw=none, legend columns=-1},
     legend image code/.code={%
      \fill[#1] (0cm,-0.1cm) rectangle (0.2cm,0.1cm);
    }   
  ]
  \addplot[blue,fill=blue!15!white,draw=none] coordinates{ (hate speech,0.9734) (abusive conversation, 1.0302)     };

  \addplot[blue,fill=blue!30!white,draw=none]  coordinates{ (hate speech,0.972) (abusive conversation, 0.8306
)        };

  \addplot[blue,fill=blue!45!white,draw=none]  coordinates{ (hate speech,1.0456)   (abusive conversation, 0.7223
)      };

\addplot[blue,fill=blue!65!white,draw=none]  coordinates{ (abusive conversation, 0.6946
)      };

\addplot[blue,fill=darkblue!70!white,draw=none]  coordinates{ (abusive conversation, 0.7065
)      };

\addplot[blue,fill=darkblue!100!white,draw=none]  coordinates{ (abusive conversation, 0.6782
)      };
  
  \legend{Top 3, Top 4, Top 5, Top 6, Top 7, Top 8}
  \end{axis}
\end{tikzpicture}
  \caption{Comparison of the ensemble system’s performances on two tasks.}
  \label{fig:re_ensemble}
\end{figure}

Figure~\ref{fig:re_ensemble} shows the ensemble system’s performances on the two datasets.
In the testing phase, we select the top-performing sub-models based on their validation accuracies. The ensemble system performs better on the abusive conversation dataset than on the hate speech dataset. Specifically, in the hate speech dataset, the best performance is achieved by the top 3 sub-models and the corresponding overall cross entropy loss is 0.9720. Conversely, the best overall cross entropy for the abusive conversation dataset is 0.6782, achieved with the top 8 (all) sub-models. The ensemble system is designed to simulate the process of annotation and has a large parameter size. Despite being trained on a substantially larger dataset, this method performs less effectively for the hate speech dataset. In this dataset, 20 annotators contribute, with each sample being annotated by five randomly assigned annotators, which means the five annotators for all the samples are not always the same individuals. As a result, one single sub-model may struggle to learn the specific characteristics of each annotator from the data. By contrast, in the abusive conversation dataset, there are eight annotators in total and for each sample it is clearly indicated which annotators are assigned for the annotation task. In this context, each sub-model is designed to emulate an individual annotator. Consequently, the ensemble system integrates the unique insights from each individual annotator, as represented by the sub-models.

\subsection{Instruction Tuning}

\begin{figure}
    \centering
    \begin{tikzpicture}
  \begin{axis}[
  width  = \linewidth,
  ybar,
    ylabel = Cross Entropy,
    x axis line style = { opacity = 0 },
    tickwidth         = 0pt,
    enlarge x limits=0.60,
    symbolic x coords = {hate speech, abusive conversation},
    xtick = data,
    ymin = 0,
    ymax = 1.8,
    ymajorgrids = true,
    legend style={at={(0.4,-0.15)},anchor=north,draw=none, legend columns=-1},
     legend image code/.code={%
      \fill[#1] (0cm,-0.1cm) rectangle (0.2cm,0.1cm);
    }   
  ]
  \addplot[blue,fill=blue!15!white,draw=none] coordinates{ (hate speech,1.2445
) (abusive conversation, 1.0627
)     };

  \addplot[blue,fill=blue!30!white,draw=none]  coordinates{ (hate speech,1.406
) (abusive conversation, 0.7883)        };

  \addplot[blue,fill=blue!45!white,draw=none]  coordinates{ (hate speech,1.6313
)   (abusive conversation, 0.6676
)      };

\addplot[blue,fill=blue!65!white,draw=none]  coordinates{ (abusive conversation, 0.62
)      };

\addplot[blue,fill=darkblue!70!white,draw=none]  coordinates{ (abusive conversation, 0.6219
)      };

\addplot[blue,fill=darkblue!100!white,draw=none]  coordinates{ (abusive conversation, 0.6448
)      };
  
  \legend{Top 3, Top 4, Top 5, Top 6, Top 7, Top 8}
  \end{axis}
\end{tikzpicture}
  \caption{Comparison of instruction tuning’s performances on two tasks.}
  \label{fig:re_instruction}
\end{figure}
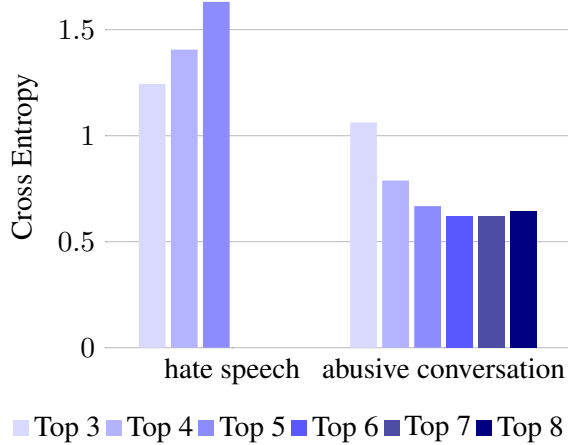

Figure~\ref{fig:re_instruction} shows the performance of the instruction tuning approach. In this approach, even though with a considerably smaller training data size, the model’s performance on the abusive conversation dataset is significantly better compared to the hate speech dataset. In the hate speech dataset, the best performance is achieved by the top 3 sub-models, with a cross entropy of 1.2445. By contrast, the lowest cross entropy in the abusive conversation dataset, achieved by the top 6 sub-models, is 0.6200. Unlike traditional machine learning or deep learning algorithms, one of the most evident advantages of instruction tuning is that it does not require much training data. Even though there are only 2501 training samples in the abusive conversation dataset, it is already sufficient to fine-tune the model and enable it to grasp the specific patterns and knowledge within the data. With this limited dataset, the pre-trained model selectively activates or deactivates certain neurons in the neural network, which serves an important role in revealing or concealing some functions embedded in LLaMa 2. Although the hate speech dataset contains a large amount of training data, the individual samples annotated by specific annotators remain unknown, which presents a challenge for the model in terms of fitting and learning patterns from the data.

\subsection{Comparison}

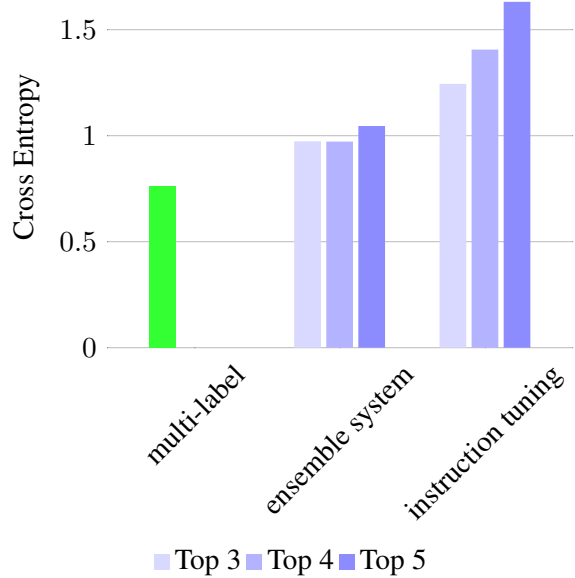
\begin{figure}
    \centering
    \begin{tikzpicture}
  \begin{axis}[
  width  = \linewidth,
  ybar,
    ylabel = Cross Entropy,
    x axis line style = { opacity = 0 },
    tickwidth         = 0pt,
    enlarge x limits=0.30,
    xticklabel style={rotate=45},
    symbolic x coords = {multi-label, ensemble system, instruction tuning},
    xtick = data,
    ymin = 0,
    ymax = 1.8,
    ymajorgrids = true,
    legend style={at={(0.4,-0.5)},anchor=north,draw=none, legend columns=-1},
     legend image code/.code={%
      \fill[#1] (0cm,-0.1cm) rectangle (0.2cm,0.1cm);
    }   
  ]

\addplot[green,fill=green!80!white,draw=none,forget plot] coordinates{ (multi-label,0.7638
)};
  
  \addplot[blue,fill=blue!15!white,draw=none] coordinates{ (ensemble system,0.9734
) (instruction tuning, 1.2445
)     };

  \addplot[blue,fill=blue!30!white,draw=none]  coordinates{ (ensemble system,0.972
) (instruction tuning, 1.406)        };

  \addplot[blue,fill=blue!45!white,draw=none]  coordinates{ (ensemble system,1.0456
)   (instruction tuning, 1.6313
)      };

  \legend{Top 3, Top 4, Top 5}
  \end{axis}
\end{tikzpicture}
  \caption{Comparison of different models’ performances on the hate speech dataset.}
  \label{fig:re_hatespeech}
\end{figure}

As shown in Figure~\ref{fig:re_hatespeech}, the multi-label method outperforms the other approaches on the hate speech dataset. The reason behind this might be the aforementioned issue in this dataset: the five annotators assigned to each sample are anonymous. Both the ensemble system and instruction tuning were trained using the same paradigm, where sub-models were fine-tuned individually on their respective labels. On the contrary, the multi-label model only relied on the probability distribution across different classes as the target, circumventing the issue with annotator anonymity. Furthermore, the hate speech dataset is big enough to fine-tune the BERT model.

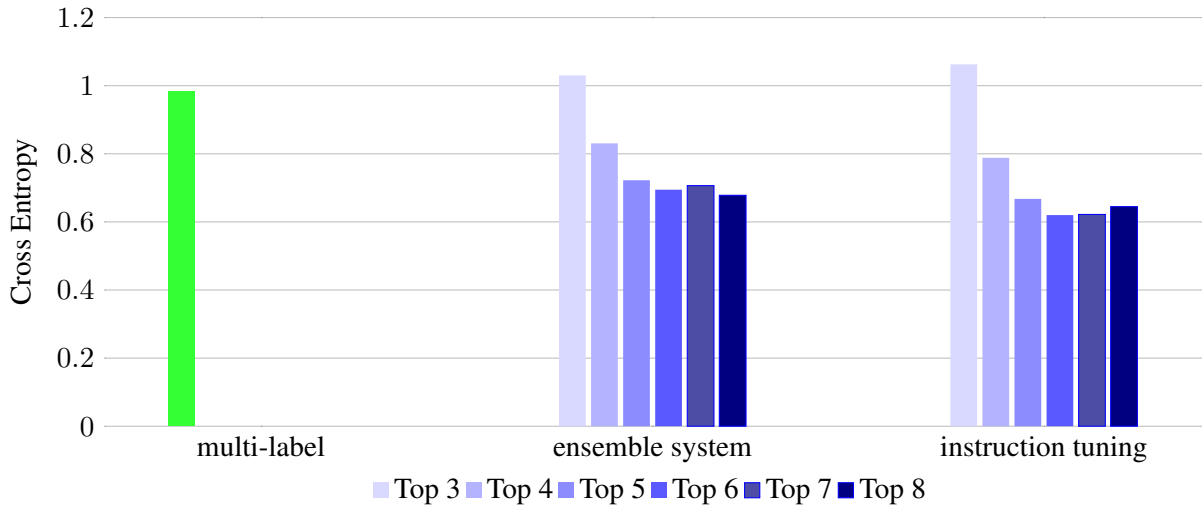
\begin{figure*}
    \centering
    \begin{tikzpicture}
  \begin{axis}[
  width  = \linewidth,
  ybar,
  height = 7cm,
    ylabel = Cross Entropy,
    x axis line style = { opacity = 0 },
    tickwidth         = 0pt,
    enlarge x limits=0.20,
    symbolic x coords = {multi-label, ensemble system, instruction tuning},
    xtick = data,
    ymin = 0,
    ymax = 1.2,
    ymajorgrids = true,
    legend style={at={(0.5,-0.1)},anchor=north,draw=none, legend columns=-1},
     legend image code/.code={%
      \fill[#1] (0cm,-0.1cm) rectangle (0.2cm,0.1cm);
    }   
  ]

\addplot[green,fill=green!80!white,draw=none,forget plot] coordinates{ (multi-label,0.9834
)};
  
  \addplot[blue,fill=blue!15!white,draw=none] coordinates{ (ensemble system,1.0302
) (instruction tuning, 1.0627
)     };

  \addplot[blue,fill=blue!30!white,draw=none]  coordinates{ (ensemble system,0.8306
) (instruction tuning, 0.7883)        };

  \addplot[blue,fill=blue!45!white,draw=none]  coordinates{ (ensemble system,0.7223
)   (instruction tuning, 0.6676
)      };

  \addplot[blue,fill=blue!65!white,draw=none] coordinates{ (ensemble system,0.6946
) (instruction tuning, 0.62
)     };

  \addplot[blue,fill=darkblue!70!white]  coordinates{ (ensemble system,0.7065
) (instruction tuning, 0.6219)        };

  \addplot[blue,fill=darkblue!100!white]  coordinates{ (ensemble system,0.6782
)   (instruction tuning, 0.6448
)      };
  
  \legend{Top 3, Top 4, Top 5, Top 6, Top 7, Top 8}
  \end{axis}
\end{tikzpicture}
  \caption{Comparison of different models’ performances on abusive conversation dataset.}
  \label{fig:re_abusive}
\end{figure*}

Figure~\ref{fig:re_abusive} shows that on the abusive conversation dataset the multi-label method performs worst. The size of this dataset is relatively small, which can result in overfitting during fine-tuning. The ensemble system consists of sub-models, with each tailored to predict annotations from a specific annotator. With multiple sub-models making their own decisions independently and contributing to the final prediction, the ensemble system can mitigate the bias brought by overfitting. Since instruction tuning does not have a high requirement for dataset size, it performs slightly better than the ensemble system.

\begin{table*}[]
\centering
\begin{tabular}{lcccccc}
\hline
\multicolumn{1}{l}{}                                                & \multicolumn{3}{c}{\textbf{Hate speech}}                 & \multicolumn{3}{c}{\textbf{Abusive conversation}}        \\ \hline
\textbf{Preference}                                                      & \textbf{Counts} & \textbf{Proportion} & \textbf{P-value} & \textbf{Counts} & \textbf{Proportion} & \textbf{P-value} \\ \hline
Baseline                                                      & 118             & 0.3278              & 0.6078           & 152             & 0.4222              & 0.0003           \\
Multi-label model                                                      & 198             & 0.5500              & \textbf{0.0000}  & 194             & 0.5389              & \textbf{0.0000}  \\
\begin{tabular}[c]{@{}c@{}}No difference\end{tabular} & 44              & 0.1222              & 1.0000           & 14              & 0.0389              & 1.0              \\ \hline
\end{tabular}
\caption{Multinomial test for probability distribution preference on two datasets.}
\label{tab:multinomial}
\end{table*}

\subsection{Online survey}
In exploring the probability distribution preference, we recruited 36 participants for the online survey. The multinomial test \cite{read_goodness--fit_2012} is employed since there are three possible preference options. The details of the results are outlined in Table~\ref{tab:multinomial}.  From this table, the multinomial tests for the multi-label model on two datasets are statistically significant, with the p-value of 0.0000. This means there is a notable disparity among the three categories being compared. Individuals generally favor the multi-label model as a more reasonable representation to characterize tweets or conversations. The results indicate the effectiveness of leveraging annotation disagreements in model training.

\section{Conclusion}
In this paper, we proposed and compared three approaches to incorporate diverse annotations in the training of ML models: a probability-based multi-label method, an ensemble system, and instruction tuning.
All three approaches take the individual labels from all annotators into account for model training in different ways, rather than only depending on an assumed “ground truth” label. In this way, the output includes a rich diversity of perspectives from annotators. We applied the proposed models on two datasets, which correspond to two tasks: hate speech detection and abuse detection in conversational AI. The two datasets show discrepancies in terms of data size, classification difficulty, the number of annotators involved in each sample, and their anonymity levels. Results show that on the hate speech dataset, the multi-label method demonstrates the highest performance among the three models, while instruction tuning achieves the lowest loss on the abusive conversation dataset.
Lastly, an online survey was conducted to evaluate the performance of the probability-based multi-label model in comparison to the baseline model.
The online survey investigated individuals’ preference between the distributions generated from the multi-label model and the baseline model. The evaluation of the survey results showed that the distribution generated from the multi-label model is considered more reasonable to characterize the texts compared to the baseline model. In the future, we would like to explore some methods or techniques to mitigate the class-imbalanced issue in the dataset. For example, there have been many popular algorithms that contribute to a relatively class-balanced dataset by over sampling \cite{chawla_smote_2002} or down sampling \cite{wilson_asymptotic_1972}. We would also like to work on investigating automatically generated prompts. Recent research has demonstrated that a concrete prompt, which consists of several discreate tokens, may not always be the most effective prompt to instruct the behavior of the model \cite{liu_gpt_2023}. Conversely, continuous embeddings of prompts, which might lack immediate human interpretability, make sense for the model itself \cite{li-liang-2021-prefix,subramani_can_2019}.  

\section*{Limitations}
There are some limitations to the experiments. Firstly, the ensemble system showed to be not suitable for the hate speech dataset, where the five annotators assigned to each sample are not fixed. In this dataset, each set of annotations used for training a sub-model can comprise annotations from multiple individuals. As a result, it becomes impossible for the sub-models to capture the specific characteristics of each annotator embedded in the annotations.

Secondly, both datasets in this study suffer from class-imbalanced problem, which can have an adverse impact on model training.
When trained on a class-imbalanced dataset, the model primarily focuses on the samples from the majority class and neglect those from the minority class, as that is an efficient strategy for minimizing the training loss. Another limitation is the inconsistency among annotators, which can introduce noise into the dataset and weaken model performance. Since our dataset lacked identifiable annotators, it was not possible to model individual annotator bias or assess inter-annotator agreement comprehensively. This constrains our ability to account for subjective variations in labeling.

Thirdly, we only leverage manually created prompts, which may introduce subjectivity and bias based on the prompt maker’s perspective \cite{tian_soft-prompt_2023}. It has been proved that manually created prompts suffer from a high degree of instability and a minor change in the prompt can result in substantial discrepancies in the model’ s performance \cite{liu_gpt_2023}.





\bibliography{anthology,custom}
\bibliographystyle{acl_natbib}




\end{document}